\documentclass[letterpaper, 10 pt, conference]{ieeeconf}  % Comment this line out if you need a4paper

\usepackage{xcolor}
\IEEEoverridecommandlockouts                              % This command is only needed if 
\usepackage{graphics,cite} % for pdf, bitmapped graphics files
\usepackage{epsfig} % for postscript graphics files
\usepackage{amsmath,amssymb,amsfonts,bbm}
\usepackage{tikz}
\usetikzlibrary{shapes,arrows}
\usetikzlibrary{arrows.meta,
                positioning}
\usepackage{algorithm}
\usepackage{algorithmic}

\usepackage[utf8]{inputenc} % allow utf-8 input
\usepackage[T1]{fontenc}    % use 8-bit T1 fonts

\DeclareMathOperator{\diag}{diag}

\newcommand{\Tau}{\mathrm{T}}

\newtheorem{definition}{Definition}

\newtheorem{remark}{Remark}

\title{\LARGE \bf
Model-Free $\mu$ Synthesis via
Adversarial Reinforcement Learning 
}

\author{Darioush Keivan$^{*1}$, Aaron Havens$^{*2}$, Peter Seiler$^3$, Geir  Dullerud$^{1}$, and Bin Hu$^{2}$% <-this % stops a space
\thanks{*Equal Contribution}% <-this % stops a space
\thanks{$^{1}$Darioush Keivan and Geir Dullerud are with the Coordinated Science Laboratory (CSL) and the Department of Mechanical Science \&  Engineering, University of Illinois at Urbana-Champaign, 
        {\tt\small \{dk12,
dullerud\}@illinois.edu}}%
\thanks{$^{2}$Aaron Havens and Bin Hu are with the Coordinated Science Laboratory (CSL) and the Department of Electrical and Computer Engineering, University of Illinois at Urbana-Champaign,
        {\tt\small \{ahavens2,
binhu7\}@illinois.edu
}}%
 \thanks{$^{3}$Peter Seiler is  with  the Department of Electrical Engineering and Computer Science, University of Michigan,
        {\tt\small pseiler@umich.edu}}%       
}

\begin{document}

\maketitle
\thispagestyle{empty}
\pagestyle{empty}

%%%%%%%%%%%%%%%%%%%%%%%%%%%%%%%%%%%%%%%%%%%%%%%%%%%%%%%%%%%%%%%%%%%%%%%%%%%%%%%%
\begin{abstract}
Motivated by the recent empirical success of policy-based reinforcement learning (RL), there has been a research trend studying the performance of policy-based RL methods on standard control benchmark problems.  
In this paper, we examine the effectiveness of policy-based RL methods on an important robust control problem, namely $\mu$ synthesis. 
We build a connection between robust adversarial RL and $\mu$ synthesis, and develop a model-free version of the well-known $DK$-iteration for solving state-feedback $\mu$ synthesis with static $D$-scaling. 
In the proposed algorithm, the $K$ step mimics the classical central path algorithm via incorporating a recently-developed double-loop adversarial RL method as a subroutine, and the $D$ step is based on model-free finite difference approximation.
Extensive numerical study is also presented to demonstrate the utility of our proposed model-free algorithm. Our study sheds new light on the connections between adversarial RL and robust control.
\end{abstract}

%%%%%%%%%%%%%%%%%%%%%%%%%%%%%%%%%%%%%%%%%%%%%%%%%%%%%%%%%%%%%%%%%%%%%%%%%%%%%%%%
\section{INTRODUCTION}

Recently, policy-based reinforcement learning (RL)~\cite{sutton2018reinforcement,schulman2015trust,schulman2017proximal} has achieved impressive performance on various control tasks~\cite{schulman2015high,lillicrap2015continuous, duan2016benchmarking}. 
Despite the empirical successes, how to choose and tune policy-based RL methods for a specific control problem at hand is not fully understood~\cite{henderson2018deep, rajeswaran2017towards}.
This inspires an increasing interest in  understanding the performance of policy-based RL algorithms on simplified linear control
benchmarks. For standard linear quadratic control problems,  policy-based RL methods have been proved to yield strong convergence guarantees in various settings \cite{pmlr-v80-fazel18a, malik2019derivative,krauth2019finite,bu2019lqr,fatkhullin2020optimizing,yang2019global,furieri2020learning,jansch2020convergence,janschporto2020policy,gravell2019learning,mohammadi2020linear,mohammadi2021convergence,matni2019self}. 
For robust/risk-sensitive control problems, the robust adversarial reinforcement learning (RARL) framework appears to be quite relevant.
An important issue for deploying RL into real-world applications is the simulation-to-real gap.
Originally RARL was developed to account for this
gap  by jointly training a protagonist and an adversary, where the protagonist learns to robustly perform
the control tasks under the possible disturbances generated by its adversary \cite{morimoto2005robust,pinto2017robust}.
Recently,
the connections between policy-based RARL and robust/risk-sensitive control have been formally studied, and policy-based RARL methods relying on double-loop update rules  have been developed to solve the $\mathcal{H}_2/\mathcal{H}_\infty$ mixed design problem and the Linear Exponential Quadratic Gaussian problem in  a provable manner \cite{zhang2020stability,zhang2021derivative,zhang2021policy}. 

An important robust control problem whose connection with policy-based RARL has been overlooked in the past is $\mu$-synthesis whose objective is to design a controller optimizing the so-called structured singular value (or equivalently the robust performance) \cite{packard93musyn}.  Over the past decade, $\mu$ synthesis has found numerous applications in industry, e.g. for robust control of hard disk drives for cloud storage \cite{honda16PhD,honda14ACC_HDD}. Reexamining the performance of RARL on $\mu$ synthesis is an important task which can lead to valuable insights regarding the connections between RL and robust control.

In this paper, we bridge the gap between policy-based RARL and state-feedback $\mu$-synthesis with static $D$-scaling.
We build upon the double-loop RARL algorithm in \cite{zhang2020stability} to develop a model-free policy optimization method for solving the state-feedback $\mu$ synthesis problem. Our proposed algorithm can be viewed as a model-free version of the well-known $DK$-iteration. In our algorithm, the $K$-step is a policy-based model-free variant of the well-established central path algorithm, and relies on the use of the double-loop RARL algorithm as the main subroutine.  The $D$-step is based on model-free finite difference approximation~\cite{conn2009introduction}. Similar to $DK$-iteration, our proposed method alternates between the $K$ and $D$ steps.
When the $D$ scaling is fixed, state-feedback $\mu$ synthesis reduces to $\mathcal{H}_\infty$ state-feedback design, and our algorithm can also be directly applied. 
The effectiveness of the proposed RARL approach on model-free  $\mu$ synthesis are demonstrated via an extensive numerical study.   Our paper complements existing work on data-driven robust control \cite{holicki2020controller,van2022,berberich2020combining} by establishing a
a new connection between model-free adversarial RL and $\mu$-synthesis. Our paper also brings new insights for understanding robust RL in general.

The rest of the paper is organized as follows. The problem formulation and some preliminaries are given in Section \ref{sec:PbF}. Next, the proposed algorithm is presented in Section \ref{sec:main}.
Then, we provide a numerical study (Section \ref{sec:numerical}) as well as some concluding remarks (Section \ref{sec:con}).

\section{Problem Formulation and Preliminaries} \label{sec:PbF}
\subsection{Notation}
{Let $\ell^m_2$ be the vector space of square-summable 
sequences in $\mathbb{R}^m$; namely, an element $x\in\ell^m_2$ is of the form $x=(x_0, x_1, x_2, \ldots )$, where each $x_k\in \mathbb{R}^m$, and its associated $\ell_2$-norm is denoted by $\|x\|_2$.}  We will frequently suppress the dependence on $m$ when clear. For a linear time-invariant (LTI) system $G$, we denote its $\mathcal{H}_\infty$ norm (or equivalently \emph{induced} $\ell_2$ norm) by $\|G\|_\infty$. 
% We say a linear mapping $M:\ell \rightarrow \ell$  is \emph{causal} if 
% for each $N\geq 1$ we have $P_N M (I-J_NP_N)=0$.  We denote the set of all causal linear mappings from $\ell^m$ to $\ell^n$  by $\mathcal{T}(\ell^m, \ell^n)$, and abbreviate to $\mathcal{T}(\ell^m)$ when $n=m$. }
\subsection{Problem Statement}
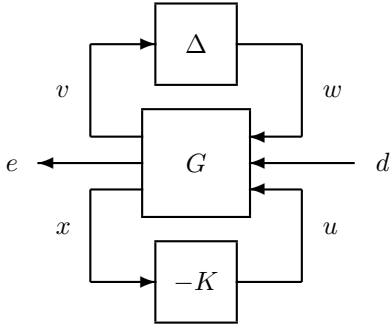
\begin{figure}[t]
\centering
\scalebox{1.0}{
\begin{picture}(172,129)(23,-15)
%\begin{picture}(172,120)(23,-15)
 \thicklines
 \put(75,25){\framebox(40,40){$G$}}
 \put(163,42){$d$}
 \put(155,45){\vector(-1,0){40}}  
 \put(23,42){$e$}
 \put(75,45){\vector(-1,0){40}}  
% I/O for Delta
% \put(80,75){\dashbox(30,30){$\Delta$}}
 \put(80,75){\framebox(30,30){$\Delta$}}
 \put(42,70){$v$}
 \put(55,55){\line(1,0){20}}  
 \put(55,55){\line(0,1){35}}  
 \put(55,90){\vector(1,0){25}}  
 \put(143,70){$w$}
 \put(135,90){\line(-1,0){25}}  
 \put(135,55){\line(0,1){35}}  
 \put(135,55){\vector(-1,0){20}}  
% I/O for Multiplier Filter
 % \put(65,90){\line(0,1){43}}
 % \put(65,133){\vector(1,0){75}}
 % \put(120,90){\line(0,1){27}}
 % \put(120,117){\vector(1,0){20}}
 % \put(140,110){\framebox(30,30){$\Psi$}}
 % \put(185,138){$v_\lambda$}
 % \put(170,133){\vector(1,0){20}}
 % \put(185,122){$w_\lambda$}
 % \put(170,117){\vector(1,0){20}}
% I/O for controller
 \put(80,-15){\framebox(30,30){$-K$}}
 \put(42,18){$x$}
 \put(55,35){\line(1,0){20}}  
 \put(55,0){\line(0,1){35}}  
 \put(55,0){\vector(1,0){25}}  
 \put(143,18){$u$}
 \put(135,0){\line(-1,0){25}}  
 \put(135,0){\line(0,1){35}}  
 \put(135,35){\vector(-1,0){20}}  
\end{picture}
} % End scalebox
\caption{Interconnection for Robust Synthesis}
\label{fig:musyn}
\end{figure}

In this section, we formulate the model-free, state-feedback $\mu$ synthesis problem and clarify the ``black-box" simulator needed in such a data-driven setting. To motivate our formulation, consider
a discrete-time robust synthesis problem as shown in Figure \ref{fig:musyn}\footnote{To be consistent with the current RL literature, we set $u_k=-K x_k$.}. The LTI system $G$ is governed by the following discrete-time state-space model:
\begin{align}
\begin{split}
\label{eq:Gmodel}
    % x_{k+1}&=A\, x_k+B_w \, w_k+B_d \, d_k+B_u \, u_k\\
    % v_k&=C_v \, x_k+ D_{wv} \, w_k+D_{dv} \, d_k+ D_{uv} \, u_k\\
    % e_k&= C_e \, x_k+ D_{we} \, w_k+D_{de} \, d_k+ D_{ue} \, u_k
    % \end{split}
    x_{k+1}&=A\, x_k+B_w \, w_k+B_d \, d_k+B_u \, u_k\\
    % \begin{bmatrix}
    % v_k\\ e_k
    % \end{bmatrix}
    % &=
    % \begin{bmatrix}
    % C_v \\ C_e
    % \end{bmatrix} x_k
    % +
    % \begin{bmatrix}
    % D_{uv}\\D_{ue}
    % \end{bmatrix}
    % u_k
    v_k&=C_v \, x_k + D_{uv} \, u_k\\
    e_k&= C_e \, x_k+ D_{ue} \, u_k
    \end{split}
\end{align}
We assume that the state of $G$ can be directly measured and a static state-feedback controller is used, i.e. $u_k= - K x_k$. In this paper, we assume that all disturbance feedthrough terms are zero, however this assumptions may be possible to relax via a computationally heavy transformation. For convenient reference, we will use $\mathcal{F}_l(G,K) $ to denote the feedback interconnection of $G$ and $K$. Thus, $\mathcal{F}_l(G,K)$ is a mapping from the input $(w,d)$ to the output $(v,e)$.

The pair $(v,w)$ satisfies $w=\Delta v$ where
$\Delta$ is a mapping in a {\emph{cone} $\mathbf{\Delta}$ of structured bounded linear operators. We call $\mathbf{\Delta}$ the uncertainty set.} Details on this general interconnection for an uncertain feedback system can be found in \cite{dullerud2013course,zhou1996robust}.
The closed-loop for a given state-feedback depends on $\Delta$ and is denoted $T_{d\mapsto e}(\Delta).$
The state-feedback will be designed to optimize the robust performance of the closed-loop as formally defined next.

\begin{definition}
The controller $K$ achieves \emph{Robust Performance of level $\gamma$} if for all $\Delta \in \mathbf{\Delta}$ satisfying $\|\Delta\|_\infty \le \frac{1}{\gamma}$, the
closed-loop is: (i) well-posed, (ii) stable, and (iii) has the mapping from $d$ to $e$ satisfying
$\|T_{d\mapsto e}(\Delta)\|_\infty \leq \gamma$. {We define $\mu_K$ to be the infimum of all such $\gamma$.  }
%induced $l_2$ gain from $d$ to $e$ that is \geir{less than or equal to} $\gamma$. 
\end{definition}

{Verifying robust performance is, in general, a fundamentally difficult non-convex problem, and accordingly so is computing $\mu_K$. Hence one typically focuses on computing an upper bound. Specifically, define a set of scaling matrices $\mathbf{D}$ with the property that for each $D\in \mathbf{D}$ we have $D \Delta = \Delta D$
for all $\Delta\in \mathbf{\Delta}$.} For a fixed controller $K$, an upper bound on  $\mu_K$ is given by the following optimization:
\begin{align}
\label{eq:rpub}
    \bar{\mu}_K = \inf_{D\in\mathbf{D}}
      \| \diag(D,I) \, \mathcal{F}_l(G,K) \, \diag(D^{-1},I) \|_{\infty} 
\end{align}
This is the so-called $D$-scale upper bound on the robust performance metric \cite{dullerud2013course,zhou1996robust,packard93}; {when  time-varying uncertainties are considered the set $\mathbf{D}$ contains only static matrices and $\bar{\mu}_K=\mu_K$.
Although more general frequency-dependent scalings can be used for LTI uncertainties,
 our paper will focus on the static diagonal $D$-scaling case for simplicity.

Figure~\ref{fig:DK} shows a block diagram representation of the scaled system that appears in the robust performance upper bound \eqref{eq:rpub}. {The goal for $\mu$ synthesis is to minimize the function $\mu_K$ over all stabilizing controllers $K$.  We denote this optimal value of $\mu_K$ as $\mu^\star$. An approach to this problem is to work with the above upper bound, and related set of $D$-scales, to minimize the induced $\ell_2$ gain from $(\tilde{w},d)$ to $(\tilde{v},e)$.} Formally, the resulting synthesis is stated as the following optimization problem:
\begin{equation} 
\label{eq:1}
 \bar{\mu}^\star = \inf_{D\in\mathbf{D},K } \| \diag(D,I) \,  \mathcal{F}_l(G,K) \, \diag(D^{-1},I) \|_\infty 	
\end{equation}
Note that $\mu^\star  =\bar{\mu}^\star$ for our specific problem when $\mathbf{\Delta}$ is the set of LTV uncertainty,  but generally $\mu^\star  \leq\bar{\mu}^\star$.
 Since we consider the state-feedback with static $D$ scaling, the problem can be reformulated as a convex program \cite{packard1991collection}. One issue is that this convex approach cannot be directly applied in the model-free setting.
An alternative approach is the so-called $DK$-iteration which alternates between optimizing over $D$ (with $K$ fixed) and optimizing over $K$ (with $D$ fixed). While each step  is convex, the alternation does not necessarily yield the global optimum for the joint optimization over $D$ and $K$. 
However, such a heuristic approach can find good solutions in many practical scenario, and we will generalize this method to the model-free setting.

\begin{figure}[t!]
\centering
\scalebox{1.0}{
\begin{picture}(152,90)(23,-10)
 \thicklines
 \put(75,25){\framebox(40,50){$G$}}
 \put(30,54){\framebox(25,20){$D$}}
  \put(135,54){\framebox(25,20){$D^{-1}$}}
 \put(193,42){$d$}
 \put(185,45){\vector(-1,0){70}}  
 \put(-7,42){$e$}
 \put(75,45){\vector(-1,0){70}}  
% I/O for Delta
% \put(80,75){\dashbox(30,30){$\Delta$}}
 %\put(80,75){\framebox(30,30){$\Delta$}}
 %\put(42,70){$v$}
 \put(75,63){\vector(-1,0){20}} 
  \put(-7,63){$\tilde{v}$}
  \put(193,63){$\tilde{w}$}
   \put(122,69){$w$}
     \put(62,69){$v$}
  \put(30,63){\vector(-1,0){25}} 
%\put(55,55){\line(0,1){35}}  
 %\put(55,90){\vector(1,0){25}}  
 %\put(143,70){$w$}
 %\put(135,90){\line(-1,0){25}}  
 %\put(135,55){\line(0,1){35}}  
 \put(135,63){\vector(-1,0){20}}  
  \put(185,63){\vector(-1,0){25}} 
% I/O for Multiplier Filter
 % \put(65,90){\line(0,1){43}}
 % \put(65,133){\vector(1,0){75}}
 % \put(120,90){\line(0,1){27}}
 % \put(120,117){\vector(1,0){20}}
 % \put(140,110){\framebox(30,30){$\Psi$}}
 % \put(185,138){$v_\lambda$}
 % \put(170,133){\vector(1,0){20}}
 % \put(185,122){$w_\lambda$}
 % \put(170,117){\vector(1,0){20}}
% I/O for controller
 \put(80,-15){\framebox(30,30){$-K$}}
 \put(42,18){$x$}
 \put(55,35){\line(1,0){20}}  
 \put(55,0){\line(0,1){35}}  
 \put(55,0){\vector(1,0){25}}  
 \put(143,18){$u$}
 \put(135,0){\line(-1,0){25}}  
 \put(135,0){\line(0,1){35}}  
 \put(135,35){\vector(-1,0){20}}  
\end{picture}
} % End scalebox
\caption{Interconnection for DK Formulation}
\label{fig:DK}
\end{figure}
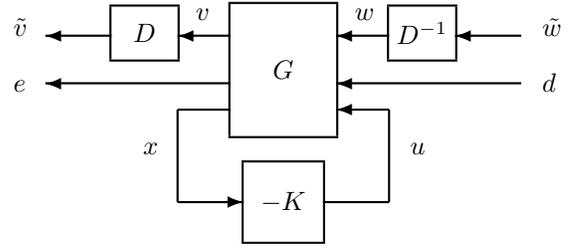

The focus of this paper is the model-free setting where all the state/input/output matrices in \eqref{eq:Gmodel} are unknown.  We only assume the availability of  a "black-box" simulator for $G$. Notice that the nominal control design corresponds to $\Delta=0$. In this case, one only needs a simulator which is capable of generating the trajectories of $\{e_k\}$ given $w=0$ and any sequence $\{d_k\}$. However, solving the robust synthesis in a model-free manner requires a more powerful simulator. We assume that the simulator for \eqref{eq:Gmodel} is able to generate the trajectories of $\{e_k, v_k,x_k\}$ for any given $\{d_k, w_k, u_k\}$.
Notice that a black box simulator for the nominal model (with no uncertainty) can be modified to incorporate $(w,v)$ channels corresponding to input multiplicative uncertainty. This is a standard uncertainty class that accounts for non-parametric error (unmodeled dynamics) at the plant input
\cite{dullerud2013course,zhou1996robust,packard93}. 
We will also demonstrate such simulator via the setting in Section~\ref{sec:numerical}.
Then the goal of our paper is to use the above ``black-box" simulator of the uncertain plant to solve the state-feedback $\mu$ synthesis problem \eqref{eq:1} with static $D$-scaling in a model-free manner.

\subsection{Model-Free Minimum-Entropy $\mathcal{H}_\infty$ Control via RARL}\label{modelfree_game}

Our proposed model-free solution for state-feedback $\mu$ synthesis will rely on existing results on RARL for linear quadratic (LQ) games. Via the RARL framework, one can design robust policies against possible adversarial attacks by jointly training a protagonist 
and an adversary via a game formulation where the protagonist learns to robustly perform the control tasks under the possible disturbances generated by its adversary. Here, we briefly review some relevant results on LQ RARL.
Consider a two-player, zero-sum, LQ game:
\begin{align}
\label{eq:Jdef}
  J(u,h) & := \mathbb{E}\sum_{k=0}^\infty \left\{x_k\top Qx_k+u_k^\top R^uu_k-h_k^\top R^h h_k\right\} \\
  \nonumber
  & \mbox{subject to: }  x_{k+1} = A x_k + B_u u_k + B_h h_k, \, x_0\sim \mathcal{D}
\end{align}
 where $Q$, $R^u$, and $R^h$ are positive definite matrices with compatible dimensions. The initial state $x_0$ is drawn from the distribution $\mathcal{D}$.
 In the RARL framework, the protagonist uses the ``control action'' $u$ to minimize
$J$ while the adversary uses the ``attack" $h$ to maximize $J$.   The expectation is taken over the trajectory $\{x_t\}$, and the only randomness stems from the random initial state satisfying $\mathbb{E}[x_0x_0^\top]=\Sigma_0$. 
 The goal for RARL is to solve the Nash equilibrium  of the above game and obtain a pair of control-disturbance sequences $\{u^*_t\}$ and $\{h^*_t\}$ satisfying $J(u^*,h)\le J(u^*,h^*)\le J(u, h^*)$
for any $u$ and $h$.
It is known~\cite{bacsar1995h} that the Nash Equilibrium of the above LQ game can be attained by state-feedback controllers, i.e., there exists a pair of matrices $(K^*,L^*)$, such that $u_t^*=-K^*x_t$ and $h_t^*=-L^*x_t$. Hence, it suffices to search over the \emph{stabilizing} control gain matrices $(K,L)$ (policy parameters). This leads to the following minimax problem where $J$ becomes a function of $(K,L)$:
\begin{align*}
\min_{K} \max_{L} & \,J(K,L)   \\
\nonumber
   \mbox{subject to: } & x_{t+1} = A x_t + B_u u_t + B_h h_t, \, x_0\sim \mathcal{D}\\
   \nonumber & u_t=-K x_t, h_t=-L x_t
\end{align*}
Therefore, one can apply various iterative gradient-based methods and their model-free counterparts to solve the above minimax problem. Based on \cite{zhang2020stability}, a naive application of the gradient descent ascent method may fail, and a double-loop algorithm can be used to guarantee convergence and stability. The double-loop algorithm uses an outer loop to update the protagonist's policy as $K_{n+1}=\textbf{PolicyOptimizer}(K_{n},L(K_n))$ where $L(K_n)=\arg\max_L J(K_n, L)$ is solved within an inner loop optimization subproblem with fixed $K_n$. The $\textbf{PolicyOptimizer}$ in the outer loop refers to a one-step update using either Gauss-Newton or a natural policy gradient method. Both the outer-loop and inner-loop updates can be implemented in a model-free manner.   For example, the least square policy iteration (LSPI) algorithm (which is the model-free counterpart of the Gauss-Newton method) \cite{lagoudakis2003least, krauth2019finite} can be applied to solve the inner-loop optimization and the one-step outer-loop update.

For our problem, the most important application of the above RARL method is to provide a model-free solver for the so-called minimum-entropy $\mathcal{H}_\infty$ control problem \cite{mustafa1990minimum}.\footnote{Depending on whether to include the ``$-$" sign into the definition of the entropy, some papers adopt the terminology ``maximum-entropy $\mathcal{H}_\infty$ control" to refer to the same problem \cite{glover1989derivation, glover1988state}.}
For a fixed $D$, let $\mathcal{K}_{\gamma}$ denote the set of all stabilizing controllers satisfying the close-loop $\mathcal{H}_\infty$ bound $\gamma$, i.e. $\| \diag(D,I) \, \mathcal{F}_l(G,K) \, \diag((D^{-1},I) \|_{\infty} \le \gamma$.  
The minimum-entropy control aims at solving the ``minimum entropy" center of $\mathcal{K}_{\gamma}$ for any given $\gamma$.
It is well known that the minimum entropy controller can be solved via an equivalent game formulation. Therefore, we can modify the above double-loop RARL algorithm to obtain a model-free oracle \textbf{RARLSolver}$(\gamma, K, D)$ which uses an internal iterative process initialized from $K$ to generate the minimum-entropy center of $\mathcal{K}_\gamma$ for any given $\gamma$ and $D$.
The implementation details for \textbf{RARLSolver} are presented in the appendix.

\begin{remark}
There exist other model-free solutions for LQ games \cite{al2007model}. However, the double-loop RARL algorithm in \cite{zhang2020stability} has strong stability and convergence rate guarantees. Hence we will mainly use RARL as our model-free solver for the intermediate minimum entropy control problem.
\end{remark}

\subsection{Model-free $\mathcal{H}_{\infty}$ Evaluation via Power Iteration} \label{PIM}
Before proceeding to our proposed model-free method for solving the design problem \eqref{eq:1}, it is natural to ask whether there exists a model-free oracle for evaluating the value of the objective function $|| \diag(D,I) \,  \mathcal{F}_l(G,K) \, \diag(D^{-1},I) ||_\infty$ given any $K$ and $D$.
The answer is yes. There are various methods available for the $\mathcal{H}_\infty$-norm estimation tasks \cite{muller2017stochastic,rojas2012analyzing,rallo2017data,wahlberg2010non,oomen2014iterative,tu2018approximation,muller2019gain,tu2019minimax}.  
One approach which is particularly suitable for our setting is the multi-input, multi-output (MIMO) power iteration method~\cite{oomen2013iteratively}. This
 relies on a specialized time-reversal method to estimate the $\mathcal{H}_{\infty}$ norm of an LTI  MIMO system from the spectral radius of its finite-time approximated representation. Given a black-box simulator for a stable system $\tilde{G}$,  the power iteration method provides an efficient oracle for estimating $||\tilde{G}||_\infty$ denoted as
 \begin{align}
   \|\tilde{G}\|_\infty\approx \textbf{HinfOracle}(\tilde{G}, N)  
 \end{align}
where $N$ is specified by the users.
%\textcolor{red}{AH: I wouldn't say that we are generating a matrix, but using $\underline{\tilde G}_N$ implicitly to estimate the max singular value directly through inputs and outputs $\underline y=\underline{\tilde G}_N \underline h$}. 
The \textbf{HinfOracle} uses the simulated input/output data of $\tilde{G}$ to query $\underline{\tilde{G}}_N$, which is an $N$-step finite-time approximation of $\tilde{G}$, and then
outputs a number to estimate  the following spectral radius
\begin{align*}
\bar{\sigma}(\underline{\tilde{G}}_N)
%=\sup_{\underline h\neq 0}\sqrt{\frac{\underline{h}^{\top} \underline{\tilde G}_N^{\top} \underline{\tilde G}_N \underline{h}}{\underline{h}^{\top} \underline{h}}}
=\sqrt{\lambda_{\max} \left(\underline{\tilde{G}}_N^{\top} \underline{\tilde{G}}_N\right)}.
\end{align*}
The $\mathcal{H}_\infty$-norm of $\tilde{G}$ can be recovered as:
\begin{equation}
    \|\tilde{G}\|_{\infty} = \lim_{N\rightarrow \infty} \bar{\sigma}\left(\underline{\tilde{G}}_N\right).
\end{equation}
The key step in the \textbf{HinfOracle} is that time-reversal 
is used to access the adjoint system $\underline{\tilde{G}}^{\top}_N$ from the input/output data generated by the simulator of $\tilde{G}$.
We refer the readers to ~\cite{oomen2013iteratively} for implementation details of the power iteration method.

It should be noted that the \textbf{HinfOracle} will typically generate  a \textit{lower-bound} for the $\mathcal{H}_{\infty}$ norm of the original system. 
Some relevant theory can be found in \cite{tu2018approximation}.
For our purpose, a tight upper bound is desired, and we will discuss a potential fix in the next section.

\section{MAIN ALGORITHM}
\label{sec:main}

As mentioned previously, in the case where the model $G$ is known, the 
robust synthesis problem~\eqref{eq:1} is typically solved via a coordinate-descent-type method called $DK$-iteration. For any fixed $(K,D)$, denote the objective function $\Gamma(K,D):=\| \diag(D,I) \, \mathcal{F}_l(G,K) \, \diag((D^{-1},I) \|_{\infty} $.
Then $DK$-iteration follows the update rule:
\begin{align}
\label{eq:Kstep}
K^{(n+1)}&=\arg\min_K \Gamma\left(K,D^{(n)}\right)\\
\label{eq:Dstep}
D^{(n+1)}&=\arg\min_D \Gamma\left(K^{(n+1)},D\right)
\end{align}
where the initial $D^{(0)}$ is usually chosen as $I$.
This approach alternates between the $K$-step \eqref{eq:Kstep} and $D$-step \eqref{eq:Dstep}. When the model is known, both steps can be efficiently solved as convex programs. In the model-free case, 
our proposed algorithm can be viewed as a sample-based counterpart of the $DK$-iteration method. Specifically, we will develop iterative model-free 
algorithms to solve both the $K$-step \eqref{eq:Kstep} and the $D$-step \eqref{eq:Dstep} in an approximate way.

\subsection{Overview}

 An overview summary of our proposed
approach is given in Algorithm \ref{al:DK}.
Our model-free algorithm still delineates two main steps: 1) a $K$-step which performs $\mathcal{H}_{\infty}$ synthesis for a fixed scaling $D$, and 2) a $D$-step which optimizes $\mu$ over static scaling matrix $D$ for a fixed $K$. 
The main difference is that  the exact minimization \eqref{eq:Kstep} \eqref{eq:Dstep} are replaced  with  model-free approximation updates \eqref{eq:approxK} \eqref{eq:approxD}.

\begin{itemize}
    \item $K$-step: In contrast to solving \eqref{eq:Kstep} exactly, we call the oracle 
 \textbf{Approx-Kmin} to obtain a model-free solution for the $\mathcal{H}_{\infty}$ synthesis with a fixed scaling $D^{(n)}$.  The oracle \textbf{Approx-Kmin} runs an iterative method by itself and requires an initial policy which is not explicitly needed in the original exact minimization \eqref{eq:Kstep}. At step $n$, We use the iterate $K^{(n)}$ to initialize the iterations in \textbf{Approx-Kmin} and the output of \textbf{Approx-Kmin} is used as $K^{(n+1)}$. On the conceptual level, the iterative algorithm within \textbf{Approx-Kmin} can be viewed as a model-free counterpart of the central path algorithm.
 The details for the model-free oracle \textbf{Approx-Kmin} are presented in Section~\ref{kstep}. 
 \item $D$-step: Similarly, the exact optimization \eqref{eq:Dstep} is replaced with a model-free oracle \textbf{Approx-Dmin} that runs an iterative finite-difference optimization method to optimize $D$ for a fixed $K$.  At step $n$, the finite-difference optimization in \textbf{Approx-Dmin} is initialized with $D^{(n)}$ and will generate an output $D^{(n+1)}$. Section~\ref{dstep} gives
 details for the model-free oracle \textbf{Approx-Dmin}.   
\end{itemize}
It is emphasized that both \textbf{Approx-Kmin} and \textbf{Approx-Dmin} only require the use of
 a ``black-box'' system $G$ of the structure~\eqref{eq:Gmodel} which  is able to generate the trajectories of $\{e_k,v_k\}$ given inputs $\{d_k,w_k\}$. 
 Both oracles heavily rely on the model-free $\mathcal{H}_\infty$ estimator \textbf{HinfOracle} introduced in Section~\ref{PIM} as well as some model-free iterative optimization methods and hence we need to provide effective initialization when calling them. It is also worth mentioning that Algorithm \ref{al:DK} requires an initial nominally stabilizing controller $K^{(init)}$, which can also be obtained using standard policy-based RL methods \cite{lamperski2020computing}. 
 We will now describe \textbf{Approx-Kmin} and \textbf{Approx-Dmin} in detail as well as some practical considerations important for implementation.

\begin{algorithm}[ht] 
\caption{Model-free $DK$-iteration}
\begin{algorithmic}[1] \label{al:DK}
\REQUIRE Simulator $G$
\STATE Input: Stabilizing $K^{(init)}$, and number of iterations $N$
\STATE \textbf{Initialize} $D^{(0)} := I$, $K^{(0)} := K^{(init)}$
% \WHILE{$DK$ termination criteria are not met} 
\FOR{$n = 0, \cdots, N-1$}
    \STATE  \textbf{Model-Free $K$-Step Section~(\ref{kstep}):} Call the oracle \textbf{Approx-Kmin}($K^{(n)},D^{(n)}$) to update $K$ as:
        \begin{align}
        \label{eq:approxK}
        K^{(n+1)} = \textbf{Approx-Kmin}(K^{(n)},D^{(n)})
        \end{align}
\vspace{-0.2in}
         \STATE  \textbf{Model-Free $D$-Step Section~(\ref{dstep}):} Call the oracle \textbf{Approx-Dmin}($K^{(n+1)},D^{(n)}$) to update $D$ as:
        \begin{align}
        \label{eq:approxD}
        D^{(n+1)} = \textbf{Approx-Dmin}(K^{(n+1)},D^{(n)})
        \end{align}
 
    \ENDFOR
\RETURN the final control design $K^{(N)}$ 
\end{algorithmic}
\end{algorithm}

\subsection{Model-free Approximation for $K$-step}\label{kstep}

Now we give details for how to solve the $K$-step in a model-free way.
The pseudo code for \textbf{Approx-Kmin} is given in Algorithm \ref{alg:model_Kstep}. The goal is to perform model-free $\mathcal{H}_\infty$ synthesis for a fixed scaling $D$. An iterative algorithm is used. 
For clarity, we use $\tilde{K}_\tau$ to denote the internal controller iterations within \textbf{Approx-Kmin}. 
When used in the $n$-th iteration of the main algorithm \ref{al:DK}, \textbf{Approx-Kmin} will initialize as $\tilde{K}_0=K^{(n)}$ and generate the final output as $K^{(n+1)}=\tilde{K}_\Tau$, where $\Tau$ is the number of the iterations run within \textbf{Approx-Kmin}.

\begin{algorithm}[!hpb] 
	\caption{Model-Free Oracle  \textbf{Approx-Kmin($K,D$)}} 
	\label{alg:model_Kstep}
	\begin{algorithmic}[1]
		\STATE Input: Initial $K$, number of iterations $\Tau$, fixed scaling $D$, finite window length $L$, scalars $\{\delta_\tau\}_{\tau=0}^{\Tau-1}$
		\STATE \textbf{Initialize}  $\tilde{K}_0:= K$
		\FOR{$\tau = 0, \cdots, \Tau-1$}
		\STATE Call \textbf{HinfOracle} to obtain an estimate for the $\mathcal{H}_\infty$ norm of the closed-loop system with controller $\tilde{K}_\tau$:
		\[
		\kern-2.8em \gamma_\tau=\textbf{HinfOracle}(\diag(D,I) \, \mathcal{F}_l(G,\tilde{K}_\tau) \, \diag((D^{-1},I),L) 
		\]
\vspace{-0.2in}
		\STATE Slightly perturb $\gamma_\tau$ and then call \textbf{RARLSolver} to solve the resultant mixed design problem:
			\[
		\tilde{K}_{\tau+1}=\textbf{RARLSolver}(\gamma_\tau+\delta_\tau, \tilde{K}_\tau, D) 
		\]
		\ENDFOR
		\STATE Return the controller  $\tilde{K}_{\Tau}$
	\end{algorithmic}
\end{algorithm}

  Next, we discuss the internal process for \textbf{Approx-Kmin}.
At each iteration $\tau$, \textbf{HinfOracle} is first called to compute the closed-loop $\mathcal{H}_\infty$ norm $\gamma_\tau$ for the associated controller $\tilde{K}_\tau$. Specifically, we want $\gamma_\tau$ to be a good estimate for $\| \diag(D,I) \, \mathcal{F}_l(G,\tilde{K}_\tau) \, \diag((D^{-1},I) \|_{\infty} $.
Let $\mathcal{K}_{\gamma}$ denote the set of all stabilizing controllers satisfying the close-loop $\mathcal{H}_\infty$ bound $\gamma$, i.e. $\Gamma(K, D)\le \gamma$. In the robust control literature, $\mathcal{K}_{\gamma}$ is also termed as the ``$\gamma$-admissible set." 
Obviously, $\tilde{K}_\tau$ is on the boundary of the set $\mathcal{K}_{\gamma_\tau}$. Intuitively, the center of $\mathcal{K}_{\gamma_\tau}$ should have a closed-loop $\mathcal{H}_\infty$ norm being smaller than $\gamma_\tau$.
If we move the iterations towards the center of 
$\mathcal{K}_{\gamma_\tau}$, we should be able to get a controller with a smaller closed-loop $\mathcal{H}_\infty$ norm and then improve the robust performance. 
This motivates our next step which 
is to call the RARL algorithm \textbf{RARLSolver} to approximately solve the ``minimum-entropy" center for the $\gamma_\tau$-admissible set. One technical subtlety is that running \textbf{RARLSolver}$(\gamma,K)$ requires the initial point $K$ to have a closed-loop $\mathcal{H}_\infty$ norm which is strictly smaller than $\gamma$. Therefore, we cannot  run \textbf{RARLSolver}$(\gamma_\tau,\tilde{K}_\tau)$ directly. We need to slightly perturb $\gamma_\tau$ to enlarge the admissible set such that $\tilde{K}_\tau$ becomes an interior point that is good for the initialization purpose. Then we can call \textbf{RARLSolver}$(\gamma_\tau+\delta_\tau,\tilde{K}_\tau)$ to generate the central solution of the set 
$\mathcal{K}_{\gamma_\tau+\delta_\tau}$. 
The perturbation parameter $\delta_\tau$ needs to be tuned in a case-by-case manner, and can be typically chosen as a small positive number as long as the estimate from \textbf{HinfNorm} is reasonable.
Then the center of $\mathcal{K}_{\gamma_\tau+\delta_\tau}$ should be close to the center of 
$\mathcal{K}_{\gamma_\tau}$.
Then we will just set the updated controller $\tilde{K}_{\tau+1}$ to be the ``central" solution generated by \textbf{RARLSolver}.
As $\tau$ increases, the set $\mathcal{K}_{\gamma_\tau}$ is expected to shrink. For sufficiently large $\Tau$, the set $\mathcal{K}_{\gamma_\Tau}$ becomes sufficiently small, and the algorithm will return a controller $\tilde{K}_{\Tau}$ which approximates the solution for the original $K$-step minimization problem~\eqref{eq:Kstep}. 

\textbf{Connections with central path:} There is a deep connection between the proposed method
\textbf{Approx-Kmin} and the well-known central path algorithm for solving semi-definite programs (SDPs)~\cite{boyd1994linear,boyd1993GEVP}. The central path algorithm computes the \textit{analytic center} $p^*(\lambda)$ of a $\lambda$-cost sublevel set of the feasible LMI through a barrier function formulation. At each iteration, $\lambda$ is updated as the cost achieved by the previous analytic center. 
Our proposed method is similar in the sense that at each iteration, $\tilde{K}_{\tau+1}$ is generated to approximate the center of the previous admissible set. Formally speaking, \textbf{RARLSolver} generates an approximate solution for the so-called minimum entropy $\mathcal{H}_\infty$ control problem, and this naturally leads to the ``central solution"  for the given $\mathcal{H}_\infty$-constrained set \cite{glover1989derivation}. 
Therefore, on the conceptual level,  \textbf{Approx-Kmin} can be viewed as an approximate version of the central path algorithm.
It is well known that
the central path algorithm yields strong convergence guarantees and is typically much faster than non-smooth methods based on the subgradient of the $\mathcal{H}_\infty$ norm.
We suspect that \textbf{Approx-Kmin} will also yield strong convergence guarantees and leave such theoretical study as future work. Notice the entropy itself is not a barrier function for the $\mathcal{H}_\infty$-constrained set, and hence new theoretical arguments may be needed. Inspired by a variant of the central path algorithm, we can also apply a factor $\theta \in [0,1)$ to interpolate $\lambda_{\tau+1} = (1-\theta)\gamma_{\tau+1} + \theta \lambda_{\tau}$. This interpolation further promotes feasibility of \textbf{RARLSolver}.

\subsection{Model-Free Approximation for $D$-step}\label{dstep}
We now develop the model-free $D$-step procedure \textbf{Approx-Dmin} with pseudo code given in Algorithm~\ref{alg:model_Dstep}. 
In the $D$-step, we fix the controller $K$ and optimize~\eqref{eq:1} over static diagonal $D$ scaling matrices to further reduce the upper-bound of the robust performance through a subgradient descent method.
Similarly to \textbf{Approx-Kmin}, we denote the internal $D$ scale iterations by $\tilde D_{\tau}$ and their parameters by $\tilde d_{\tau}$.
At the $n$-th iteration of our main Algorithm~\ref{al:DK}, we apply \textbf{Approx-Dmin}, with $\tilde D_{0} := D^{(n)}$, to generate the update $D^{(n+1)} := \tilde{D}_{\Tau}$.

\begin{algorithm}[!hpb] 
	\caption{Model-Free Oracle  \textbf{Approx-Dmin($K,D$)}} 
	\label{alg:model_Dstep}
	\begin{algorithmic}[1]
		\STATE Input: Fixed $K$, initial parameterized scaling $D=\diag(\exp(d))$, number of iterations $\Tau$, finite window length $L$, gradient step-size $\alpha$, perturbation size $\varepsilon$.
		\STATE \textbf{Initialize}  $\tilde{d}_0 := d$
		\FOR{$\tau = 0, \cdots, \Tau-1$}
		\STATE Use the \textbf{CentralDiffOracle} to estimate the central difference gradient update for the $\mathcal{H}_\infty$ norm of the closed-loop system with controller $K$ and scaling elements $\tilde{d}_\tau$:
		\[
		\kern-2.8em \tilde{d}_{\tau+1} = \textbf{CentralDiffOracle}(\tilde d_{\tau}, \varepsilon, \alpha, L)
		\]
\vspace{-0.2in}
		\ENDFOR
		\STATE $\tilde{D}_{\Tau} := \diag(\exp(\tilde{d}_{\Tau}))$
		\STATE Return the scaling matrix  $\tilde{D}_{\Tau}$
%\[ 
%\widehat{\nabla_K
%  C(K,L)} = \frac{1}{m} \sum_{i=1}^m \frac{d}{r^2} \widehat C_i U_i
%\, , \quad
%\widehat \Sigma_{K,L} = \frac{1}{m} \sum_{i=1}^m \widehat \Sigma_i .
%\]
	\end{algorithmic}
\end{algorithm}

Gradient descent is well-suited since the objective~\eqref{eq:1} is differentiable almost everywhere with respect to $D$ and can be made convex through an exponential parameterization of $D$~\cite{safonov1984minimizing}.
In the model-free setting, we can only query evaluations of the objective function~\eqref{eq:1} through our data-driven power method \textbf{HinfOracle}, so we employ the finite difference method to obtain subgradient estimates. The static diagonal $D$ scaling is explicitly parameterized by $D:=\exp(\diag(d))$, where $d = (d_1,\ldots,d_m)$ and $m$ is the dimension of the diagonal uncertainty $\Delta$. We now define the convex function $H$ as
\begin{equation}
    H(d) := ||\diag(e^{\diag(d)},I) \mathcal{F}_l(G,K)\diag(e^{-\diag(d)},I)||_{\infty}
\end{equation}
The \textit{central difference} estimate of the $j$th entry of the gradient $g$ is
\begin{equation}
      g_j(d,\varepsilon) = \frac{H(d + \varepsilon e_j) -H(d-\varepsilon e_j)}{2\varepsilon},\quad j\in [m]
\end{equation}
% \begin{equation}
%      g_j = \frac{h(D+\varepsilon E_j) - h(D-\varepsilon E_j)}{2\varepsilon}
% \end{equation}
%Where $E_j$ is the matrix whose $(j,j)$-th element is $1$ and all other entries are $0$ \textcolor{red}{AH: Maybe someone knows a better way to denote this perturbation} 
where $e_j$ is a vector who's $j$th entry is $1$ and zero elsewhere and $\varepsilon$ is some positive number sufficiently small. With the gradient $g$ in hand, we can proceed to perform gradient descent updates on $d$. Using the black-box simulator and \textbf{HinfOracle} with a window of $N$ to take evaluations of $H$, we have an approximate oracle for the finite difference gradient update on our internal parameter $\tilde d_\tau$:
\begin{align*}
  \tilde d_{\tau+1} = \tilde d_{\tau} - \alpha g(\tilde d_\tau,\varepsilon)\approx \textbf{CentralDiffOracle}(\tilde d_\tau, \varepsilon, \alpha ,N)
\end{align*}
Where \textbf{CentralDiffOracle} is applied for a fixed number of iterations and the final scaling is given by $\tilde D_{\Tau} := \diag(\exp(\tilde d_\tau))$. Note that when using unreliable evaluations, $\varepsilon$ must be chosen large enough to observe a valid descent direction, but also small enough to capture the local gradient information and avoid oscillations around a minimum.
Notice that we use the above proposed finite difference method manly for its simplicity. There are more sophisticated nonsmooth optimization methods \cite{burke2020gradient,kiwiel2010nonderivative} which can be used to improve the convergence for the above $D$-step calculation.

\section{NUMERICAL CASE STUDY}
%\section{NUMERICAL EXAMPLE: UNCERTAIN COUPLED SPRING-MASS SYSTEM}
\label{sec:numerical}
In order to demonstrate the effectiveness of our proposed model-free $DK$-iteration algorithm~\ref{al:DK}, we present a numerical example of a MIMO system with input-multiplicative uncertainty. We will compare our model-free procedure to a standard model-based $DK$ synthesis method to make empirical observations of accuracy and convergence characteristics.

\textbf{Uncertain Coupled Spring-Mass System: } 
%\subsection{Uncertain Coupled Spring-Mass System} 
The nominal system is a fully-observable, coupled mass-spring system with two control inputs and four outputs. The model is extended to include two input uncertainties and two disturbances which enter through the same channel as the control input (i.e. $B_u=B_w=B_d$). The continuous-time state-space matrices  and outputs for this system are given by:
\begin{align*}
    A &= 
    \begin{bmatrix}
    0 & 0 & 1 & 0\\
    0 & 0 & 0 & 1\\
    -k/m_1 & k/m_1 & 0 & 0\\
    k/m_2 & -k/m_2 & 0 & 0 
    \end{bmatrix}, B_{u} =
    \begin{bmatrix}
    0 & 0\\ 0 & 0\\ 1/m_1 & 0\\ 0 & 1/m_2
    \end{bmatrix}\notag\\
    C_v &= 0,\quad C_e =\begin{bmatrix}
    0\\ Q^{1/2}
    \end{bmatrix},\quad
    D_{uv} = I,\quad D_{ue} =\begin{bmatrix}
    R^{1/2}\\0
    \end{bmatrix}
\end{align*}
with $m_1=1.0$, $m_2=0.5$, $k=1.0$, 
$Q=I$ and $R=0.1I$.

%$Q^{1/2}=I$ and $R^{1/2}=\sqrt{0.1}I$.

\begin{figure}[t!]
\centering
\scalebox{1.0}{
\begin{picture}(145,119)(23,20)
 \thicklines
 
\put(205,63){\vector(1,0){20}} 
\put(212,69){$e_x$}
\put(180,53){\framebox(25,20){$Q^{\frac{1}{2}}$}}
\put(160,63){\line(1,0){20}} 
\put(168,69){$x$}
\put(120,40.5){\framebox(40,45){$\tilde{G}$}}
\put(25,76){\vector(1,0){78}}
\put(107,76){\vector(1,0){13}}
\put(27,68){$u_1$}
\put(25,50){\vector(1,0){78}}
\put(107,50){\vector(1,0){13}}
\put(27,42){$u_2$}
\put(50,76){\line(0,1){15}}
\put(105,78){\line(0,1){13}}
\put(105,76){\circle{4}}
\put(105,67){\vector(0,1){7}}
\put(75,83.5){\framebox(20,15){$\Delta_1$}}
\put(50,91){\vector(1,0){10}}
\put(60,86){\framebox(10,10){$\alpha$}}
\put(70,91){\line(1,0){5}}
\put(40,82){$\tilde{v}_1$}
\put(95,91){\vector(1,0){10}}
\put(106,82){$\tilde{w}_1$}
\put(108,66){$d_1$}
\put(50,50){\line(0,1){15}}
\put(105,52){\line(0,1){13}}
\put(105,50){\circle{4}}
\put(105,36){\vector(0,1){12}}
\put(50,65){\vector(1,0){10}}
\put(40,56){$\tilde{v}_2$}
\put(95,65){\vector(1,0){10}}
\put(106,56){$\tilde{w}_2$}
\put(108,35){$d_2$}
\put(75,57.5){\framebox(20,15){$\Delta_2$}}
\put(60,60){\framebox(10,10){$\alpha$}}
\put(70,65){\line(1,0){5}}

\put(30,76){\line(0,1){48}}
\put(37.5,50){\line(0,1){66}}
\put(30,124){\vector(1,0){30}}
\put(37.5,116){\vector(1,0){22.5}}
\put(60,110){\framebox(25,20){$R^{\frac{1}{2}}$}}
\put(85,120){\vector(1,0){20}}
\put(92,126){$e_u$}

\put(-11,45){\framebox(36,36){$-K$}}
\put(-26,63){\vector(1,0){15}}
\put(-26,63){\line(0,-1){40}}
\put(-26,23){\line(1,0){196}}
\put(170,63){\line(0,-1){40}}

\end{picture}
} % End scalebox
\caption{Closed loop interconnection for a fully-observable coupled mass-spring system with input multiplicative uncertainty. The scaled plant $\tilde{G}$ is the interconnection of plant $G$ and two scaled matrices $D$ and $D^{-1}$. Where $D$ is a $2 \times 2$ diagonal matrix.  $\tilde{v}_1 = u_1$ and $\tilde{v}_2 = u_2$ are inputs of one dimensional uncertainties $\Delta_1$ and $\Delta_2$, respectively. Uncertainties outputs $\tilde{w}_1 = \Delta_1 \tilde{v}_1$ and $\tilde{w}_2 = \Delta_2 \tilde{v}_2$ along with control inputs $u_1$, $u_2$ and disturbance inputs $d_1$ and $d_2$ enter scaled plant $\tilde{G}$ as the plant inputs . $\tilde{w}:=(\tilde{w}_1^{\top},\tilde{w}_2^{\top})^{\top}$, $d:=(d_1^{\top},d_2^{\top})^{\top}$, $u:=(u_1^{\top},u_2^{\top})^{\top}$ and $\tilde{v}:=(\tilde{v}_1^{\top},\tilde{v}_2^{\top})^{\top}$ and $e:=(e_u^{\top},e_x^{\top})^{\top}$ are system inputs and outputs in Figure \ref{fig:DK}, respectively.}
\label{fig:uncertain_block_diagram}
\end{figure}
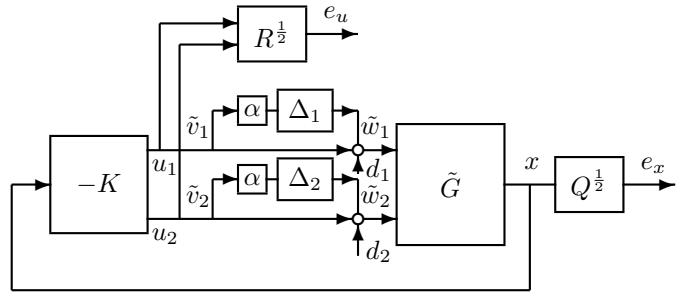

 The discrete-time matrices are given by a zero-order-hold discretization of $A$ and $B_u$ with a sample time $t_s=0.1$. 
 As in the $\mu$ synthesis problem~\eqref{eq:1}, we aim to minimize the gain from $h:=(\tilde w, d)$ to $(\tilde v, e)$, where $e:=(e_u^{\top},e_x^{\top})^{\top}$ is a performance output describing the state error $e_x$ and control effort $e_u$. The multiplicative uncertainty is at the plant input as seen in Figure~\ref{fig:uncertain_block_diagram}, which can account for a standard class of non-parametric errors. It is standard  to design for normalized uncertainty $||\Delta||_{\infty}\leq 1$ (i.e. to aim for $\bar \mu \leq 1$). We include a constant factor of $\alpha=0.25$ at the uncertainty input. This corresponds to an effective uncertainty of 25\% in each input channel.  We emphasize that our approach is model-free and does not require knowing the above model parameters.

Using \eqref{eq:cost} in the appendix, we obtain the following dynamic game cost for a fixed $\gamma$ and diagonal scaling $D$:
\begin{align}
    J(u,h) = \mathbb{E}\sum^{\infty}_{k=1} \{x_k^{\top}Qx_k &+ u_k^{\top}(R+D^2)u_k\notag\\
    &- \gamma^2 h_k^{\top} \diag(D^2, I) h_k\}.
\end{align}
The $K$-step is solved with this cost
using the model-free \textbf{RARLSolver}.
In this case study, the procedure \textbf{Approx-Dmin} is applied over $2\times 2$ diagonal scaling matrices parameterized by $D = \exp(\diag(d_1, d_2))$. The \textbf{Approx-Kmin}~(algorithm \ref{alg:model_Kstep}) is implemented using LSPI to perform the \textbf{RARLSolver} subroutine. An error constant of $\delta_0=1\mathrm{e}{-1}$ and $\delta_\tau=5\mathrm{e}{-3}$ for all $\tau>0$ is added to each new $\gamma_\tau$ estimate given by \textbf{HinfOracle}. The first constant is made to be larger because the first closed-loop under $\tilde K_{0}$ is often less robust and \textbf{HinfOracle} typically yields a higher error. We continue to apply the \textbf{RARLSolver} subroutine until $\gamma$ can no longer be decreased by a small threshold value, where we then proceed to apply \textbf{Approx-Dmin} once again. 

Figure~\ref{fig:DK_experiment} shows
results from five total $DK$-iterations. We compare the values of $\bar\mu$ found by our Model-free $DK$-iteration algorithm~\ref{al:DK} and fully model-based $DK$-iteration.\footnote{Notice that the model-based $DK$-iteration and the convex approach in \cite{packard1991collection} generate almost the same results on this example.}
The horizontal axis of Figure~\ref{fig:DK_experiment} is the number updates on $K$ that occur within the \textbf{RARLSolver} subroutine. \textbf{RARLSolver} will continue to update $K$ until convergence, where it returns $\tilde{K}_{\tau+1}$ and a new $\gamma_\tau$ is computed, hence the piece-wise form of the graph. The iteration at which \textbf{Approx-Dmin} is applied is denoted by grey lines. As we get closer to the optimal $\bar \mu$ upper-bound, convergence in \textbf{Approx-Kmin} occurs more quickly and \textbf{Approx-Dmin} occurs more frequently.

In this first $K$-step when $D=I$, the application of \textbf{Approx-Kmin} is equivalent to $\mathcal{H}_{\infty}$ synthesis of the uncertain loop in Figure~\ref{fig:uncertain_block_diagram}. The model-free \textbf{Approx-Kmin} approaches the optimal nominal $\mathcal{H}_{\infty}$ value when $D=I$. Moreover, the minimum upper bound $\bar \mu$ after Model-free $DK$-iterations~(algorithm~\ref{al:DK}) approaches within $2\%$ of the value found through model-based $DK$-iterations. Ultimately, the level of accuracy is determined by the accuracy of \textbf{HinfOracle} and the size of the error constants $\{\delta_\tau\}_{\tau}$ which must be large enough to yield an upper-bound. Clearly there is a trade-off between the number of samples expended on \textbf{HinfOracle} and the accuracy of $\bar\mu$. Fortunately, using a window length of $L=100$ was sufficient to achieve the performance shown in Figure~\ref{fig:DK_experiment}.   Although there is no theoretic convergence guarantee for our proposed algorithm, this case study provides evidence that our algorithm is capable of  achieving a near optimal $\gamma$ in a model-free manner for the case of $\mathcal{H}_{\infty}$ synthesis and $\mu$-synthesis. As shown in Figure~\ref{fig:DK_experiment}, the convergence requires running the $K$-step for roughly $2500$ times. Each time requires a full trajectory and consumes about $0.6$ seconds of CPU time on a laptop.  Hence the total CPU time  is roughly $25$ minutes. 
Of course this is much slower than model-based approaches. However,
we emphasize that our goal is not to propose a computationally efficient method outperforming existing approaches in the model-based setting. It is our hope that the connection between adversarial RL and $\mu$-synthesis can shed new light on data-driven robust control.
\begin{figure}[t!]
    \centering
    \includegraphics[width=\linewidth]{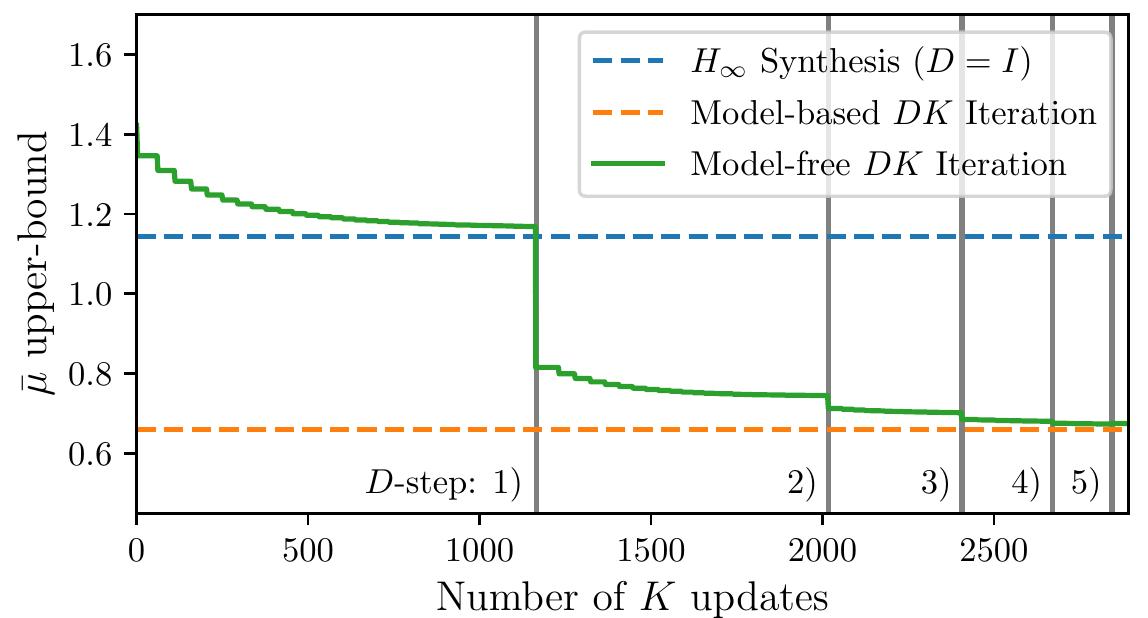}
    \caption{The model-free $DK$-iteration Algorithm~\ref{al:DK} is initialized with a stabilizing controller and five $DK$-iterations are run. For each update on $K$ within \textbf{RARLSolver}, we plot the true achieved upper-bound $\bar \mu$ of problem~\eqref{eq:1}. The value is piece-wise constant since a new $\bar \mu$ is only determined after completing the \textbf{RARLSolver} subroutine. The gray lines denote when a $D$ step occurs, hence the instantaneous jumps. Values of $\bar \mu$ are also shown for nominal model-based $\mathcal{H}_{\infty}$ synthesis and $DK$-iteration. 
    %The value found by the model-free $DK$ algorithm~\ref{al:DK} differs from the model-based by less than $2\%$.
    }
    \label{fig:DK_experiment}
\end{figure}

\section{CONCLUSIONS}
\label{sec:con}
 
In this paper, we examine the effectiveness of policy-based RARL methods on the robust $\mu$ synthesis problem. 
We have developed a model-free version of the well-known $DK$-iteration for solving state-feedback $\mu$ synthesis with static D-scaling. 
In the proposed algorithm, the $K$ step can be viewed as an approximate version of the central path algorithm where centers are solved via a recently-developed double-loop RARL algorithm.  The $D$ step is based on model-free finite difference approximation.
When the D scaling is fixed, $\mu$ synthesis reduces to $\mathcal{H}_\infty$ synthesis, and our algorithm can also be directly applied. Numerical evidence is provided to demonstrate the effectiveness of the proposed method. It is our hope that our study can inspire more research on connecting adversarial RL and modern robust control.

\section*{ACKNOWLEDGMENT}
D. Keivan and G. Dullerud are partially funded by NSF under the grant ECCS 19-32735.
A. Havens and B. Hu are generously supported by the NSF award CAREER-2048168 and the 2020 Amazon research award.
P. Seiler is supported by the 
US ONR grant N00014-18-1-2209.

\bibliography{bibtex}
\bibliographystyle{IEEEtran}

%\begin{thebibliography}{0}

%\end{thebibliography}

\section*{APPENDIX}%\label{appendix:append_A}
%\subsection{LSPI Implementation for RARL}

We present more details for the implementation of \textbf{RARLSolver}.
For our general system structure $G$ (governed by the LTI model~\eqref{eq:Gmodel}) with static diagonal scaling matrix $D$, the $\mathcal{H}_{\infty}$ synthesis problem minimizes the gain from $\{(\tilde w,d)\}$ to $\{(\tilde v,e)\}$ as shown in the Figure ~\ref{fig:DK} with the state space form:
\begin{equation}
\begin{cases}
x_{k+1} = A x_k + B_u u_k +\begin{bmatrix} B_w D^{-1}  & B_d  \end{bmatrix} \begin{bmatrix} \tilde w_k \\ d_k \end{bmatrix} &
\\
\begin{bmatrix}
        \tilde v_k \\ e_k
\end{bmatrix} = \begin{bmatrix}
        D C_v \\ C_e
\end{bmatrix} x_t + \begin{bmatrix}
        D D_{uv} \\ D_{ue}
\end{bmatrix} u_k &
        
\end{cases}
\end{equation}

First, we must translate the above setting to an LQ dynamic game introduced in section~\ref{modelfree_game} using only the original \textit{unscaled} simulator $G$ (i.e. using data inputs $\{(u_k, d_k, \tilde w_k)\}^N_{k=0}$ and outputs $\{(x_k, e_k, v_k\}^N_{k=0}$). The resultant minimum entropy $\mathcal{H}_\infty$ control problem can be formulated as a game~\cite{bacsar1995h}:

\begin{align}\label{eq:cost}
    %c_k(x_k, u_k, h_k) 
    &J(u,h) \nonumber\\  =&\mathbb{E} \sum^{\infty}_{k=0} \{e_k^{\top}e_k + v_k^{\top}D^2v_k - \gamma^2  h_k^{\top} \diag(D^2, I) h_k\}\nonumber\\
=& \mathbb{E}\sum^{\infty}_{k=0}\{ x_k^{\top} Q x_k + u_k^{\top} R_u u_k+ 2 x_k^{\top}N u_k - h_k^{\top} R_v h_k\}
\end{align}
where $h = (\tilde w^{\top}, d^{\top})^{\top}$ and the cost matrices can be given explicitly in terms of the state space matrices of $G$.

\begin{align}
    Q &= C_v^{\top} D^2 C_v + C_e^{\top}C_e,&&R_u = D_{uv}^{\top}D^2 D_{uv} + D_{ue}^{\top}D_{ue}\nonumber\\
    N &= C_v^{\top} D^2 D_{uv}+C_e^{\top} D_{ue},&&R_v =\gamma^2 \diag(D^2,I)
\end{align}
The double-loop algorithm can be applied to the above shifted system to give the oracle \textbf{RARLSolver}.
Notice that we needed to introduce crossing terms $N$ which RARL algorithm can still address such without difficulty. 

Within \textbf{RARLSolver}, a double-loop algorithm is used. At each round $n$, we first fix $\hat{K}_n$ and run an inner-loop iteration to maximize \eqref{eq:cost} by choosing $L$. 
When $\hat{K}_n$ is fixed, the system dynamics become
$x_{k+1} = (A-B_u \hat{K}_{n})x_{k} + \begin{bmatrix} B_w & B_d  \end{bmatrix} h_{k}$.
		Maximizing \eqref{eq:cost} subject to such dynamics
leads to an indefinite LQR problem which can be efficiently solved by LSPI within the inner loop. 
It is also possible to apply other RL methods such as REINFORCE or natural policy gradient to solve the resultant inner-loop indefinite LQR problem. In this paper, we use the LSPI method which has better empirical performance for the LQR problem \cite{krauth2019finite}. 

Once the inner-loop problem is solved, we obtain an attacker's policy $L(\hat{K}_n)$ which is the worst-case attacker's policy for the fixed $\hat{K}_n$. Next, we fix this attacker's policy and do a one-step LSPI update in the outer-loop. The system becomes
$x_{k+1} = (A-\begin{bmatrix} B_w & B_d \end{bmatrix} L(\hat{K}_n) )x_{k} + B_{u} u_{k}$. 
Minimizing \eqref{eq:cost} subject to such dynamics
leads to another LQR problem, and we can apply the LSPI algorithm to make a one-step update.
By repeating such iterations, the \textbf{RARLSolver} is able to find the Nash equilibrium of  the game and give  the minimum-entropy center of the $\gamma$-admissible set.

Notice that LSPI can be implemented in an off-policyline manner. So for each inner loop, we only need to sample one set of data and then we can iterate the LSPI update on the same data.
However, for each outer loop, we also need to resample data for the fixed $L(\hat{K}_n)$. This increases the sample complexity.
Finally, it is worth mentioning that LSPI has no difficulty in handling the crossing terms in our cost function, although existing theory does not directly address this case.

\end{document}